\def\year{2022}\relax
\documentclass[letterpaper]{article} 
\usepackage{aaai22}  
\usepackage{times}  
\usepackage{helvet}  
\usepackage{courier}  
\usepackage[hyphens]{url}  
\usepackage{graphicx} 
\usepackage{natbib}  
\usepackage{caption} 
\DeclareCaptionStyle{ruled}{labelfont=normalfont,labelsep=colon,strut=off} 
\usepackage{algorithm}
\usepackage{algorithmic}

\usepackage{newfloat}
\usepackage{listings}
\floatstyle{ruled}
\newfloat{listing}{tb}{lst}{}
\floatname{listing}{Listing}
\usepackage{soul}
\usepackage{url}
\usepackage[hidelinks]{}
\usepackage[utf8]{inputenc}
\usepackage{caption}
\usepackage{graphicx}
\usepackage{amsmath}
\usepackage{booktabs}
\usepackage{tikz}
\tikzset{every picture/.style={remember picture}}
\usetikzlibrary{tikzmark}
\usepackage{multirow}
\usepackage{makecell}
\usepackage{hhline}
\usepackage[normalem]{ulem}
\usepackage{paralist}

\usepackage{xcolor}

\definecolor{mygreen}{RGB}{50, 128, 50}

\title{Automated Story Generation as Question-Answering}

\author{Louis Castricato, 
  Spencer Frazier,
  Jonathan Balloch, 
  Nitya Tarakad, and 
  Mark O. Riedl}
  
\affiliations{
    \texttt{\{lcastric, sfrazier7, balloch, ntarakad3, riedl\}@gatech.edu}
}

\usepackage{bibentry}

\newcommand\sysname[1]{{\sc EDGAR}}

\begin{document}

\maketitle

\begin{abstract}
Neural language model-based approaches to automated story generation suffer from two important limitations. 
First, language model-based story generators generally do not work toward a given goal or ending.
Second, they often lose coherence as the story gets longer.
We propose a novel approach to automated story generation that treats the problem as one of generative question-answering. 
Our proposed story generation system starts with sentences encapsulating the final event of the story.
The system then iteratively (1)~analyzes the text describing the most recent event, (2)~generates a question about ``why'' a character is doing the thing they are doing in the event, and then (3)~attempts to generate another, preceding event that answers this question.
%
We evaluate the coherence of generated stories using human participant studies that show that stories generated by our system are 15\% more coherent ($p<0.013$) than those generated by BART fine-tuned on a story corpus to generate backward from a given ending. 
\end{abstract}

\section{Introduction}


Consider a story, at the ending of which a princess is reunited with her lover thought to be lost at sea, a swordsman has enacted revenge on the man who killed his father, and a giant becomes a pirate. 
One might reasonably wonder how this situation came to pass. 
Aristotle writes in {\em Poetics} that the events of the story serve the plot and the end.
Under this interpretation, storytelling is explanation---every event answers the question of ``how did the next event come to pass?''.
In this paper, we propose an automated story generation system using the principles of question-answering and show how it can improve automated story generation capabilities.  

Automated Story Generation is the challenge of designing an artificial intelligence system that can generate a story from a minimal number of inputs---often just a prompt and some storytelling knowledge.
Symbolic story and plot generation systems have traditionally relied on planning
or case-based reasoning
(see \citeauthor{Gervs2009ComputationalAT} \shortcite{Gervs2009ComputationalAT} for an overview of symbolic story generation systems).
Some of these systems start with an end state---the state the fictional world should be in at the end of the story---and work backward, determining what must have happened to transform an initial world state into the goal.
These systems often generate coherent stories guaranteed to end in a given state.
Their drawback is that they require significant  hand-authored domain knowledge.

Machine learning-based story generation systems
acquire or learn story domain knowledge from data, often corpora of human-authored stories. 
Most machine learning-based story generation systems have relied on neural network-based language models. 
Auto-regressive neural language models trained on a corpus of stories learn a probability distribution over tokens $p(t_n|t_{n-1}, t_{n-2}, ...,t_{n-k})$ based on the tokens that occur in the training corpus.
This distribution can then be sampled to create new texts that emulate the training corpus.
Training a neural language model on story corpora results in a generative model that produces texts that look like stories~\cite{roemmele2016writing,khalifa2017deeptingle,martin2018event}.
However, language model based approaches are unable to bring stories to a particular conclusion or goal state.
Stories generated by language models also tend to lose coherence over time as they rely on probabilistic sampling and do not learn a richer model of the story world.

We consider how neural story generation systems can be induced to generate more coherent narratives that also end in a pre-determined, desirable way.
Narratives are perceived to be coherent when events are related to each other in a way that is comprehensible by the reader~\cite{trabasso1985causal,graesser91}.
There are many relations between events which fit this need, the most important are: (1) causal relations---one event cannot happen if another event had not happened prior to it---and (2) character goal hierarchies---an action is in service of a goal or another action that is in service of a goal.

Our insight is that if each event in the story is generated to explicitly answer the question of ``why'' the next event in the story happens, then readers will perceive the story as more coherent.
To generate a story that will be perceived as a coherent and build up to a pre-determined ending, we propose to generate the story backward. 
This is achieved by starting from a textual description of the final event; each event added best answering the question of what must have proceeded it.
Our system, \sysname{}, repeats this process for a specified number of iterations.
Questions are generated using a commonsense inference model, Para-COMET~\cite{gabriel2020paragraph}, to predict what readers are likely to believe about a story event; the inferences are transformed into questions using templates.
\sysname{} then attempts to answer each question using a generative question-answering model.

We evaluate our system against a baseline neural transformer-based language model approach that is fine-tuned to generate story events backward, matching the backward process of \sysname.
We measure story coherence with two human-participant studies. 
In the first, perceived coherence is measured as the entropy in participant responses to true/false questions about the story~\cite{CastricatoFabula2021,Castricato2021Formal}; a story that is more comprehensible results in less random guessing by human readers.
We find that \sysname{} generates more coherent stories than the baseline as evidenced by the entropy of answers about stories generated by \sysname{} had 15.9\% lower entropy than those of the baseline. 
The second evaluation is subjective---we qualitatively measure coherency via subjective questionnaire about coherence. 
Participants consider stories written by \sysname{} twice as coherent as those written by the baseline.


\section{Related Work}

\citeauthor{Gervs2009ComputationalAT} \shortcite{Gervs2009ComputationalAT} overviews early symbolic story generation systems.
Story generation systems that use symbolic story planners utilize logic-like domain representations that provide knowledge about available actions, their preconditions, and their effects.
A search process---such as that by \citeauthor{riedl2010narrative}~\shortcite{riedl2010narrative}---selects a goal condition or a precondition of an action in the plan and attempts to find another, preceding action that has an effect that establishes the condition. 
This process iterates, creating chains of preconditions and effects until everything is grounded in the initial world state.
However, the chaining can be done forward from the initial state to the goal as well~\cite{ware2010modeling,ware2021}.

Neural networks have the potential to generate a greater range of stories by learning model for how to tell stories from a corpus of exemplar stories.
Neural language models learn the probability that one or more tokens will occur given a history of one or more prior tokens, $P_{\theta}(t_{n+1}, ..., t_{n+m}| t_{n-k}, ..., t_{n-1}, t_{n})$, according to token occurrence patterns in a corpus.
Neural language models can be induced to generate text that can be read as a story by sampling from the learned distribution over tokens and appending them to a prompt. 
Some neural language model based story generation techniques include~\cite{roemmele2016writing,martin2018event,khalifa2017deeptingle}. 
However, a neural language model alone is incapable of achieving a specific end state or event.
Sampling from a distribution over tokens only considers the most likely successive tokens given a window of prior tokens.
Neural language models also tend to lose story coherence over time.
This is due to the fact that a language model only models a distribution over tokens in the training set. Additionally, the hidden parameters of current neural networks are unlikely to encode the state of a fictional world, as human readers would understand.

\citeauthor{tambwekar2018controllable}~\shortcite{tambwekar2018controllable} attempt to train a neural language model to generate toward a given goal.
They fine-tune a neural language model with a policy-gradient reinforcement learning technique that rewards the language model for generating events progressively closer to the goal event.
This has the benefit of improving readers' perceptions of coherence, but---being based on a language model---does not ensure that any transition from one event to the next will always be perceived as related.

Other neural language model approaches to story generation using neural networks use {\em hierarchical conditioning}, in which a high-level guidance specification is given either periodically or per sentence in the story~\cite{fan2018hierarchical,rashkin2020plotmachines,ammanabrolu2020}.
These high-level guidance specifications turn the generation problem into a supervised learning problem. 
We do not consider these approaches further in this paper because we do not assume the existence of a guidance specification.
A high level plan can be generated~\cite{yao2019plan,fan2019strategies,ammanabrolu2020}, which elevates the challenges of maintaining coherence to a higher level of abstraction but cannot reliably achieve a given story ending.

Our work is most similar to that of \citet{PengInferring2021}, which uses a commonsense inference model to match inferences at time step $t-1$ to those at time $t$.
In contrast our approach works backwards and uses a commonsense inference model to generate a ``why'' question about a character, which is then used to prompt generation of story events that are preceding.
The C2PO system~\cite{ammanabrolu2020automated} also uses the COMET~\cite{bosselut2019comet} commonsense inference engine.
However, this work uses COMET to generate successor and predecessor events, performing a bi-directional search from a given start event and a given end event.
It is relevant to our work in that it does partially chain backward from a given end event, and also uses a commonsense inference engine.
However, C2PO generates plots made up of short statements of character intentions, whereas our system generates stories with more natural-looking language.



\section{The \sysname{} System}

\begin{figure*}[t]
\centering
\includegraphics[width=\textwidth]{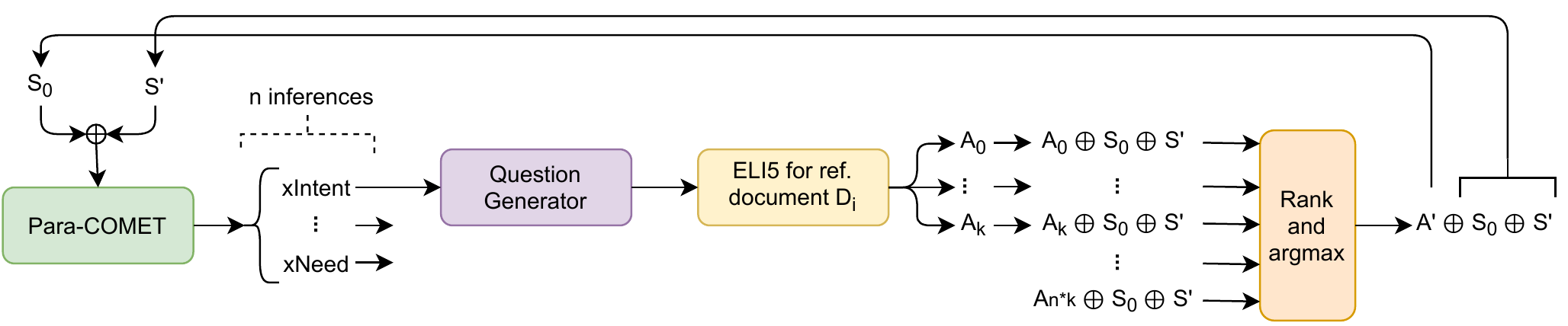}
\caption{\sysname{} generates stories backward. Given the end of the story where $S_0$ is the earliest event sequence and $S'$ is the remainder, Para-COMET generates a set of $n$ inferences. Each inference is converted a question and the ELI5 QA model generates $k+1$ answers. The answers are concatenated to the beginning of the story and the ranker selects the best scoring story.
This process is repeated.}
\label{fig:pipeline}
\end{figure*}

The {\em Explanatory Drama Generation And Recall} (\sysname{}) system constructs a story backwards from a given sentence describing the end of the story. 
%
The system contains three major components. The first component is a question generator. 
Given a story context---the sequence of text describing the earliest event in the ending context---a set of questions about the event is generated. 
Second, a question answering component attempts to generate text describing one or more events that answer that question.
A number of candidate answers are generated for each question.
Finally, the answers are iteratively pre-pended to the context and a ranker chooses the best sequence. 
The best sequence is added to the story and the process iterates. 
See the pipeline in Figure~\ref{fig:pipeline}.

\subsection{Question Generation}

We use Para-COMET~\cite{gabriel2020paragraph} to generate questions.
Para-COMET is a commonsense inference model trained to generate potential commonsense explanations about one or more sentences of input text.
Inferences made by Para-COMET have types.
In particular, \texttt{xIntent} explains what a character in the sentences might have intended in performing any actions in the sentences.
These correspond to goal relations in reader comprehension~\cite{graesser91}.
\texttt{xNeed} explains what a character might have needed to perform any actions in the sentences.
These provide precondition-like inferences, corresponding to causal relations in reader comprehension~\cite{trabasso1985causal}. 
We discard all other relation types.
Because Para-COMET works on multi-sentence sequences, we extract a rolling window of the last 5 \texttt{xIntent} and \texttt{xNeed} inferences.

Unfortunately, Para-COMET does not identify which character is associated with each \texttt{xIntent} and \texttt{xNeed}, which is problematic for stories with more than one character. 
To associate the \texttt{xNeed} and \texttt{xIntent} clauses with a character, we generate the following templates: 

\begin{itemize}
    \item ``Who needs to $xIntent$''
    \item ``Who needs $xNeed$''
\end{itemize}
filling in the details of the inferences.
These filled templates are provided as input to RoBERTa~\cite{liu2019roberta}, a question-answering model that, when provided the context and the template will output the names of the characters most likely to have had these needs and intents.

Finally, we use a second set of templates to assemble the final set of questions:
\begin{itemize}
    \item ``Why does $character$ do $xIntent$?''
    \item ``What does $character$ do to need $xNeed$?''
\end{itemize}
%
This process generates a total of 8 questions.

\subsection{Question Answering}

Once we have a set of questions, \sysname{} generates candidate answers, such that each candidate can be added to the beginning of the story context so far. 
To generate sentences describing the preceding event that answers the questions generated, we feed the questions into the ELI5 QA model~\cite{fan2019eli5}. 
The ELI5 QA model is a long-form, question-answering model trained on the {\em Explain Like I'm Five} Reddit corpus,\footnote{\url{https://www.reddit.com/r/explainlikeimfive/}} in which people give long, yet easily comprehensible answers to open-ended questions as one might give to a five-year old.
ELI5 QA requires a reference document from which to abstract answers.
%
The reference document is the source material---in this case a story---that ELI5 QA uses to generate an answer.
Because \sysname{} is an unsupervised technique designed to generate novel stories, there is no one reference document that should be used; using a single reference document would run the risk of accidentally recreating a human-written story. 
For every iteration we randomly select a reference document from the {\em Flash Fiction Online} repository.\footnote{\url{https://www.flashfictiononline.com}} 
The question templates above were constructed to induce relatively short answers from ELI5, which has a tendency to generate very long explanations.

We use beam search to generate 15 candidate answers for each question.
As another measure to prevent ELI5 from providing overly verbose explanations, we have accumulated a list of over 700 unique banned phrases, which occur when ELI5 commentators point out ``facts'' or likening a character's action to mental disability. This blocked phrases list was accumulated iteratively, by rerunning the model repeatedly and adding any toxic phrases to this excluded list.
The result is $n\times k$ story continuations where $n$ is the number of questions, $k$ is the number of beams per question on ELI5.



\subsection{Ranking}

Once \sysname{} has generated a set of candidates, the final step in the process is to select the best candidate for pre-pending to the context (the end of the story).
We prepend each answer to the context  and rate each resulting text sequence using GPT-2~\cite{radford2019language} to assess the probability of the sequence.
GPT2 was fine-tuned on the science fiction summary corpus~\cite{ammanabrolu2020} dataset, which consists of 2,276 high-quality plot summaries from science fiction TV and movie wikis. We fine-tune on the science fiction summary corpus because wiki plots do not include descriptive details or dialogue; our ranker thus prefers more plot-like narrative content. 
Candidates are ranked by perplexity of the GPT2 model. 
The normalized perplexity distribution over the beams outputted by ELI5 refers to the $1 - \textit{probability}$ distribution of a body of text existing within the distribution of science fiction summaries. 

Ranking is an important step because of the numerous processes involved; as a consequence the ranking of ELI5 beam distribution does not necessarily correlate with the final ranking, which roughly measures fluency.
The best scoring candidate is added to the overall story. 
The process repeats with the new, longer story, attempting to determine what happened just before the new context.



\section{Objective Evaluation}

We hypothesize that \sysname{}, by virtue of question-answering, can generate more coherent stories than a pure language modeling technique.
We define coherence as any perceivable relationship between events in a story.
Research on reading comprehension~\cite{trabasso1985causal,graesser91} suggest that causal and goal relationships are particularly important.


\begin{table*}[t]
    \centering
    \footnotesize
    \begin{tabular}{p{3.2in}||p{3.2in}}
    \multicolumn{1}{c}{\textbf{Stories generated by EDGAR}} & \multicolumn{1}{c}{\textbf{Stories generated by bBART}}\\
    \hline
He needs to find a way out of the house.
If he wants to go to the kitchen, he must first find a way to escape from the house.
This means that he must either run away or fight his way past the lock.
He goes to the bathroom.
The only way to get to the kitchen is through a locked door, which Hansel has never been able to do before.
However, when he opens the door, it doesn't take him very long at all.
\textbf{Hansel's hand still trembles as he pushes open the twice-cooked door.
The last time he saw the house he was glancing back over his shoulder as he and his sister fled into the trees.} & 
\multirow{2}{3.2in}{A young boy named Hansel has been walking through town recently looking at some strange things that are going on outside their walls. In fact, they were just playing cards before anyone noticed them.
A young boy called Hansel has Been walking through town lately looking at some odd things that are gone missing outside their walls -- maybe even someone noticing them. In reality, they were probably playing cards before everyone noticed them. He wants to learn more about it though, but can always tell for sure. He wants to know more about it, but can never tell for sure.
After all, this is a city of humans who have no interest in living here.He wants to know what happened to him, but can only tell for sure if this is a City of Humans who have little interest in living there. Hansel’s hand still feels like an old man with no interest whatsoever.  His mind wanders frantically around trying to figure out how to live here. 
\textbf{Hansel’s hand still trembles as he pushes open the twice-cooked door. The last time he saw the house he was glancing back over his shoulder as he and his sister fled into the trees.}}
\\
\hhline{-||}
If it does not have a pulse then it is dead.
The heart rate of an animal tells them if it is alive or dead.
There is no such thing as absolute death.
He is able to tell the difference between something being alive and something being dead, so when he looks at the house, he feels like he’s seeing things that aren’t there.
This makes him feel uncomfortable because he doesn’t want to be in that situation.
It’s similar to how people can see ghosts or monsters from inside their head but they don’t know what those things are.
{\bf Hansel’s hand still trembles as he pushes open the twice-cooked door.
The last time he saw the house he was glancing back over his shoulder as he and his sister fled into the trees.}&\\
\hline
    \end{tabular}
    \caption{Example stories generated by \sysname and backwards BART (bBART). 
    Bold text indicates the given ending prompt.}
    \label{tab:stories}
\end{table*}


Common automated evaluation metrics for story generation such as perplexity and BLEU are insufficient as they only measure whether a generator can recreate the ground truth corpus. A story may deviate from the ground truth and be considered a good story---indeed this is a desirable property of an automated story generator.
Furthermore, systems such as ours may be unsupervised and have many components that intentionally push a language model away from any one corpus, thus making perplexity less meaningful.
For these reasons, story generation research often relies on human participant studies with subjective questions.

We assert human participant studies are the best way to assess the coherence of generated stories.
Question-answering protocols, wherein questions are asked about a story, have been proposed as a means to make human-participant evaluations more objective~\cite{riedl2010narrative,cardona2016}.
We conduct a human-participant evaluation using a metric based on the {\em Fabula Entropy Indices}~\cite{CastricatoFabula2021}, a set of objective measures of story coherence grounded in narratological formalisms~\cite{Castricato2021Formal}.
The Fabula Entropy Indices are validated to be correlated with more common, subjective evaluation metrics, which we also replicate in this paper to provide further experimental evidence.

\subsection{Baselines}

Because \sysname{} generates stories from a given ending, the system cannot be evaluated relative to other systems that generate forward and are thus not guaranteed to reach a given goal.
The BART~\cite{lewis2019bart} neural language model was used as a baseline, but fine-tuned to generate events backward to conform to \sysname{} and guarantee the presence of a given end event. 
The dataset used to fine-tune consisted of $2276$ narratives from a science fiction summary corpus \cite{ammanabrolu2020}. The narratives are preprocessed to create our dataset. From every narrative, $2+2k$ sequential sentences are obtained, where $k$ is a random integer less than $5$. The $2+2k$ sentences are split apart into $2$ sentences and $2k$ sentences, creating the source and target of the dataset respectively. The $2$ sentences generated in the $2+2k$ sentence chunk always precede the $2k$ sentences, establishing a relationship between sequential sentences. We preprocess this data to this format because an attribute found in most narrative summaries within our dataset is that preceding sentences to any given sentences gives some notion of causality. 
%
%
BART utilizes seq2seq as its translation architecture. As a consequence of the input data format, our fine-tuned Backward-BART---which we refer to as bBART---can generate narratives backwards by assessing the causality between sequentially sentences. 

Human-written stories from the ROCStories corpus~\cite{DBLP:journals/corr/MostafazadehCHP16} were are also included in our evaluation as a point of comparison. These stories have a definitive causality between sequential sentences. 

\subsection{Method}

We detail the Fabula Entropy Index methodology we use below.
To evaluate the objective coherence of stories, we turn to cognitive psychology.
Cognitive psychology research suggests that recall is strongly correlated with narrative causal coherence~\cite{trabasso1985causal}.
The cognitive load of inferring entailments about a story is strongly correlated with how well the story conveys information about its fabula\footnote{A story's {\em fabula} denotes the chronological sequence of events in a narrative.}~\cite{carney2019}.
We devise a new evaluation methodology wherein we ask participants to read stories and then answer true/false questions about how the events of the story relate to each other.  
We measure the amount of agreement between readers' answers in terms of {\em entropy}.
If the story is coherent, readers will come to the same conclusions about the truth or falseness of the questions, and entropy will be low.
If the story is incoherent, readers---forced to choose between true and false---will choose more randomly, resulting in higher entropy.
We do not require a ground truth ``correct'' answer to each question in order to compute the entropy; this is a desirable property of our methodology given 
(1)~there are no algorithmically produced ground truth answers to the true/false questions and 
(2)~obtaining a ground truth answer from humans can be noisy.
Our index method is inspired by the evaluation used in \citeauthor{li:acs2012}~\shortcite{li:acs2012} where human participants were asked to choose event orderings and participant agreement was assessed as entropy.

We generated 11 stories using \sysname{}, 11 stories using backward-BART, and randomly selected 11 stories from the ROCStories corpus.
Stories were generated by running the respective systems 3 iterations.
Stories ranged from 5 sentences to 20 sentences in length.
See Table~\ref{tab:stories} for examples from \sysname{} and bBART.
The Appendix gives the entire set of stories used in the evaluation.
For the 33 stories, we produced 7 true/false questions for each story using the technique described in Section \ref{sec:entropy}.
To avoid bias, a non-computer-science graduate student not affiliated with the research project was paid to write the questions using the above template.

We recruited 180 human-subject participants from Mechanical Turk.
Participants were recruited from countries that speak English as the national language.
Each participant was asked to read 3 stories and answer the 7 true/false questions after each story.
Participants were paid \$7 for a 15 minute task.
The first story is a ``screener'' story, an uncommon fable that is easy to understand.
If a participant did not answer the questions how we expect, we eliminated the participant from the pool.
Participants were also eliminated from the pool if they resorted to marking all questions true or false or marked questions in some otherwise visually obvious repeating pattern; we eliminated 26 participants.

\subsection{Computing the Entropy Index}
\label{sec:entropy}

We define the entropy index of each story as follows.
For a given story generation system, we randomly selected 11 generated stories.
For each story, we then produced 7 entailment questions about each story.
Entailment questions are of the form of implications. 
By asking the reader to answer true or false we are asking the reader to prove or disprove the statement within the realm of what has been presented about the story world. 

In order to ensure our questions were not biased, we provide annotators the following templates, two of which are given as examples here:
\begin{itemize}
    \item $E_i$ depends on $E_j$
    \item $E_i$ could be removed and the story would still make sense.
\end{itemize}
$i < j$ and $E$ refers to an event within the story. 
The full set of templates can be found in the appendix.
The questions themselves were manually written to ensure grammatically correctness and readability. 

The answers to the entailment questions give us a measure of entropy. 
When participants disagree, it can be determined how ambiguous their model of the story world is, such that they must rely heavily on external bias.

Consider that we have some story, $S$, composed of an event chain $E = \{E_i\}_n$. An event chain being a sequence of events discussed in a story, one path in a fabula. Generate two events, one that could be inserted into $E$ and preserve coherence and its negation. We'll refer to these events as $A$ and $B$. Refer to their insertions as $E^A$ and $E^B$. Assume that we had some function $f(\cdot)$ that could take either $E^A$ or $E^B$ and rank all of the explanations for $A$ and $B$ respectively by mental load induced on the reader. Then, if $E^B$ is coherent, consider what mental leaps are required by the reader for justification. Let $D(A)$ and $D(B)$ refer to these normalized distributions respectively. Measure the following:
\begin{equation}
\begin{split}
&\text{KL}_A = \text{KL}(D(A) \| U) \text{ and } \text{KL}_B = \text{KL}(D(B) \| U)
\end{split}
\end{equation}
Where {\em KL} is Kullback–Leibler divergence and $U$ is a uniform distribution. 
Inductively if $A$ and $B$ are in direct contradiction of each other, we can collapse the above statement to
\begin{equation}
\text{KL}_{A,B} = \text{KL}\Big(D\Big(\sum f\big(E^A\big),\sum f\big(E^B\big)\Big)\| U\Big)
\end{equation}
In this case, since $U$ is of dimension two, simplify the above to entropy. We can conclude that measuring the coherence of such an insertion is equivalent to measuring the entropy over the answers to a similarly constructed T/F statement about a causal relationship within a story. Over a large number of questions and stories per model, the above serves as a sound proxy for coherence. Consider a coherent story and a set of T/F questions concerning this story. It is often easier to disprove a statement about a coherent story than it is to prove a statement about an incoherent story~\cite{o1992comprehension,albrecht1993updating}. 
By utilizing the format of T/F questions, the above will tend to converge to zero on a coherent story as there will always be one option that is disprovable. 
To get a large enough sample, we used 77 questions per model over 11 stories.

\subsection{Results}

\begin{figure}[t]
\centering
\includegraphics[scale=0.4]{"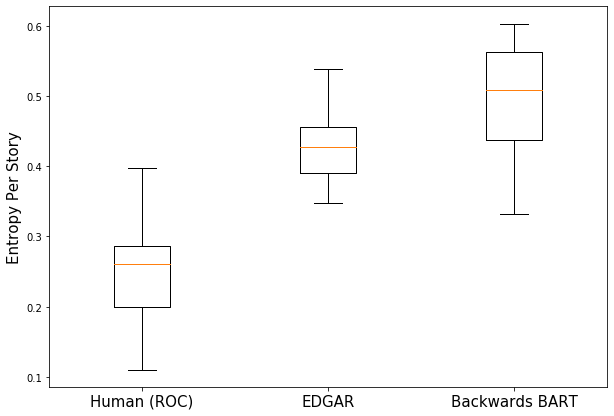"}
\caption{The entropy indices for human-written stories, \sysname{}, and backwards BART (bBART). Lower is better.
}
\label{fig:plot}
\end{figure}

The evaluation results are plotted in Figure~\ref{fig:plot}.
The evaluation shows that \sysname{} scores a median of 0.427 on the entropy index, compared to bBART's median of 0.508. 
Human written stories from the ROCStories corpus scored a median entropy of 0.26.

From these results we can draw a number of conclusions.
First, the median entropy of human-authored stories is over 95\% better than bBART and over 63\% better than EDGAR.
This implies that human-authored stories are much more coherent than computer-generated stories according to our Entropy Index metric.
This is the expected result and shows that our Entropy Index metric is operating as expected.
The human story entropy index is a lower bound.
Importantly, the median entropy \sysname{} is 15.9\% lower than that of the bBART baseline, indicating that our technique has improved the coherence of generated stories when generating backwards in order to ensure a given ending. 


\section{Subjective Evaluation}

We conducted a second human-participant evaluation in which participants read stories and answered subjective questions about the coherence of the stories.
We would expect the results of this experiment to concur with the results of the previous experiment. 

\citeauthor{purdy2018predicting}~\shortcite{purdy2018predicting} proposes a number of questions to be used to evaluate story generation systems.
They have been used in a number of story generation system evaluations (cf. \cite{tambwekar2018controllable,ammanabrolu2020,ammanabrolu2020automated,PengInferring2021}) as well as to validate the Fabula Entropy Indices.
We use a subset of the questions and adapt them to rank-order choice between stories from two systems:
\begin{itemize}
    \item Which story's events occur in a more PLAUSIBLE ORDER?
    \item Which  story's  sentences  MAKE  MORE  SENSE  given sentences before and after them?
    \item Which story better follows a SINGLE PLOT?
    \item Which story is of HIGHER QUALITY?
    \item Which story is more ENJOYABLE?
\end{itemize}
The first three questions ask about different aspects of perceived story coherence.


\begin{table}[t]
\footnotesize
\centering
\begin{tabular}{l|c|c|c}
    \hline
    {\bf Question} & {\bf EDGAR} & {\bf bBART} & {\bf $p$-value} \\
    \hline
    Plausible & 31 & 15 & 0.013 \\
    Single plot & 32 & 14 & 0.005 \\
    Makes sense & 29 & 17 & 0.052\\
    Quality & 31 & 15 & 0.013 \\
    Enjoyable & 31 & 15 & 0.013 \\
    \hline
\end{tabular}
\caption{Total counts of times per question in the subjective evaluation that participants selected a story generated by each system. $P$-tests were determined to ensure that the chance of EDGAR winning a pairing was greater than 50/50.}
\label{tab:subjective}
\end{table}

\subsection{Method}

We used the same stories from the first evaluation and the same baselines.
Participants read two stories from two different sources back-to-back.
Then for that pair of stories, the participant was asked to answer the subjective questions above, picking between the two stories.

We recruited 48 human-subject participants from Mechanical Turk.
Participants were recruited from countries that speak English as the national language.
Each participant was asked to read 4 stories, presented in pairs of two, and answer the 5 questions after each story.
Participants were paid \$5 for a 10 minute task. We screened participants by asking them similar questions about human written stories but inserted the answers to the questions in the directions, to determine their attentiveness. Participants that were considered inattentive where disqualified. 



\subsection{Results}

The results are summarized in Table~\ref{tab:subjective}, which shows the number of times, per question, a participant selected the story from each system.
When forced to pick between stories generated by \sysname{} and stories generated by Backward-BART, participants chose stories generated by \sysname{} twice as often for every question asked. 
A one-tailed binomial $p$-test for the results of each question determines \sysname{} was significantly preferred above the baseline for every dimension at $p<=0.013$ except the ``Makes sense'' dimension, which was significant at $p=0.052$.
These results suggest that \sysname{} generates more coherent and overall better quality stories than Backward-BART. 
These results are consistent with the Entropy Index metric, confirming that the metric is also measuring coherence.

\section{Conclusions}

We propose a new approach to neural story generation that treats story generation as question-answering problem---given an ending, the story must answer the question of how the ending comes about. 
Our proposed \sysname{} system generates backward from the ending event to ensure the presence of the desired ending.
It decomposes the generation process into distinct processes for using human commonsense to produce questions and then to answer them. 
These processes are grounded in reader narrative comprehension.
We show that stories generated by \sysname{} are more coherent than stories generated in a more conventional language modeling approach based on subjective and objective measures of perceived coherence.
The \sysname{} technique is a significant departure from techniques that sample from a language model that opens up new avenues for improving neural story generation in ways that are inspired by the comprehension needs of the human reader.


\bibliography{bib}

\begin{thebibliography}{32}
\providecommand{\natexlab}[1]{#1}

\bibitem[{Albrecht and O'Brien(1993)}]{albrecht1993updating}
Albrecht, J.~E.; and O'Brien, E.~J. 1993.
\newblock Updating a mental model: Maintaining both local and global coherence.
\newblock \emph{Journal of experimental psychology: Learning, memory, and
  cognition}, 19(5): 1061.

\bibitem[{Ammanabrolu et~al.(2020{\natexlab{a}})Ammanabrolu, Cheung, Broniec,
  and Riedl}]{ammanabrolu2020automated}
Ammanabrolu, P.; Cheung, W.; Broniec, W.; and Riedl, M.~O. 2020{\natexlab{a}}.
\newblock Automated Storytelling via Causal, Commonsense Plot Ordering.
\newblock \emph{arXiv preprint arXiv:2009.00829}.

\bibitem[{Ammanabrolu et~al.(2020{\natexlab{b}})Ammanabrolu, Tien, Cheung, Luo,
  Ma, Martin, and Riedl}]{ammanabrolu2020}
Ammanabrolu, P.; Tien, E.; Cheung, W.; Luo, Z.; Ma, W.; Martin, L.~J.; and
  Riedl, M.~O. 2020{\natexlab{b}}.
\newblock {Story Realization: Expanding Plot Events into Sentences}.
\newblock In \emph{AAAI Conference on Artificial Intelligence}, 8.

\bibitem[{Bosselut et~al.(2019)Bosselut, Rashkin, Sap, Malaviya, Celikyilmaz,
  and Choi}]{bosselut2019comet}
Bosselut, A.; Rashkin, H.; Sap, M.; Malaviya, C.; Celikyilmaz, A.; and Choi, Y.
  2019.
\newblock COMET: Commonsense transformers for automatic knowledge graph
  construction.
\newblock \emph{arXiv preprint arXiv:1906.05317}.

\bibitem[{Cardona-Rivera et~al.(2016)Cardona-Rivera, Price, Winer, and
  Young}]{cardona2016}
Cardona-Rivera, R.~E.; Price, T.~W.; Winer, D.~R.; and Young, R.~M. 2016.
\newblock Question Answering in the Context of StoriesGenerated by Computers.
\newblock \emph{Advances in Cognitive Systems}, 4: 227--245.

\bibitem[{Carney(2019)}]{carney2019}
Carney, J. 2019.
\newblock \emph{Necessary Fictions: Supernormal Cues, Complex Cognition, and
  the Nature of Fictional Narrative}, 391--413.
\newblock University of Nebraska Press.

\bibitem[{Castricato et~al.(2021{\natexlab{a}})Castricato, Biderman,
  Cardona-Rivera, and Thue}]{Castricato2021Formal}
Castricato, L.; Biderman, S.; Cardona-Rivera, R.~E.; and Thue, D.
  2021{\natexlab{a}}.
\newblock Towards a Formal Model of Narratives.
\newblock In \emph{Proceedings of the 3rd Workshop on Narrative Understanding}.

\bibitem[{Castricato et~al.(2021{\natexlab{b}})Castricato, Frazier, Balloch,
  and Riedl}]{CastricatoFabula2021}
Castricato, L.; Frazier, S.; Balloch, J.; and Riedl, M.~O. 2021{\natexlab{b}}.
\newblock Fabula Entropy Indexing: Objective Measures of Story Coherence.
\newblock In \emph{Proceedings of the 3rd Workshop on Narrative Understanding}.

\bibitem[{Fan et~al.(2019)Fan, Jernite, Perez, Grangier, Weston, and
  Auli}]{fan2019eli5}
Fan, A.; Jernite, Y.; Perez, E.; Grangier, D.; Weston, J.; and Auli, M. 2019.
\newblock Eli5: Long form question answering.
\newblock \emph{arXiv preprint arXiv:1907.09190}.

\bibitem[{Fan, Lewis, and Dauphin(2018)}]{fan2018hierarchical}
Fan, A.; Lewis, M.; and Dauphin, Y. 2018.
\newblock Hierarchical neural story generation.
\newblock \emph{arXiv preprint arXiv:1805.04833}.

\bibitem[{Fan, Lewis, and Dauphin(2019)}]{fan2019strategies}
Fan, A.; Lewis, M.; and Dauphin, Y. 2019.
\newblock Strategies for structuring story generation.
\newblock \emph{arXiv preprint arXiv:1902.01109}.

\bibitem[{Gabriel et~al.(2020)Gabriel, Bhagavatula, Shwartz, Bras, Forbes, and
  Choi}]{gabriel2020paragraph}
Gabriel, S.; Bhagavatula, C.; Shwartz, V.; Bras, R.~L.; Forbes, M.; and Choi,
  Y. 2020.
\newblock Paragraph-Level Commonsense Transformers with Recurrent Memory.
\newblock \emph{arXiv preprint arXiv:2010.01486}.

\bibitem[{Gerv{\'a}s(2009)}]{Gervs2009ComputationalAT}
Gerv{\'a}s, P. 2009.
\newblock Computational Approaches to Storytelling and Creativity.
\newblock \emph{AI Mag.}, 30: 49--62.

\bibitem[{Graesser, Lang, and Roberts(1991)}]{graesser91}
Graesser, A.; Lang, K.~L.; and Roberts, R.~M. 1991.
\newblock Question Answering in the Context of Stories.
\newblock \emph{Journal of Experimental Psychology: General}, 120(3): 254--277.

\bibitem[{Khalifa, Barros, and Togelius(2017)}]{khalifa2017deeptingle}
Khalifa, A.; Barros, G.~A.; and Togelius, J. 2017.
\newblock Deeptingle.
\newblock \emph{arXiv preprint arXiv:1705.03557}.

\bibitem[{Lewis et~al.(2019)Lewis, Liu, Goyal, Ghazvininejad, Mohamed, Levy,
  Stoyanov, and Zettlemoyer}]{lewis2019bart}
Lewis, M.; Liu, Y.; Goyal, N.; Ghazvininejad, M.; Mohamed, A.; Levy, O.;
  Stoyanov, V.; and Zettlemoyer, L. 2019.
\newblock Bart: Denoising sequence-to-sequence pre-training for natural
  language generation, translation, and comprehension.
\newblock \emph{arXiv preprint arXiv:1910.13461}.

\bibitem[{Li et~al.(2012)Li, Lee-Urban, Appling, and Riedl}]{li:acs2012}
Li, B.; Lee-Urban, S.; Appling, D.~S.; and Riedl, M.~O. 2012.
\newblock Crowdsourcing Narrative Intelligence.
\newblock \emph{Advances in Cognitive Systems}, 2: 25--42.

\bibitem[{Liu et~al.(2019)Liu, Ott, Goyal, Du, Joshi, Chen, Levy, Lewis,
  Zettlemoyer, and Stoyanov}]{liu2019roberta}
Liu, Y.; Ott, M.; Goyal, N.; Du, J.; Joshi, M.; Chen, D.; Levy, O.; Lewis, M.;
  Zettlemoyer, L.; and Stoyanov, V. 2019.
\newblock Roberta: A robustly optimized bert pretraining approach.
\newblock \emph{arXiv preprint arXiv:1907.11692}.

\bibitem[{Martin et~al.(2018)Martin, Ammanabrolu, Wang, Hancock, Singh,
  Harrison, and Riedl}]{martin2018event}
Martin, L.; Ammanabrolu, P.; Wang, X.; Hancock, W.; Singh, S.; Harrison, B.;
  and Riedl, M. 2018.
\newblock Event representations for automated story generation with deep neural
  nets.
\newblock In \emph{Proceedings of the AAAI Conference on Artificial
  Intelligence}, volume~32.

\bibitem[{Mostafazadeh et~al.(2016)Mostafazadeh, Chambers, He, Parikh, Batra,
  Vanderwende, Kohli, and Allen}]{DBLP:journals/corr/MostafazadehCHP16}
Mostafazadeh, N.; Chambers, N.; He, X.; Parikh, D.; Batra, D.; Vanderwende, L.;
  Kohli, P.; and Allen, J.~F. 2016.
\newblock A Corpus and Evaluation Framework for Deeper Understanding of
  Commonsense Stories.
\newblock \emph{CoRR}, abs/1604.01696.

\bibitem[{O'Brien and Albrecht(1992)}]{o1992comprehension}
O'Brien, E.~J.; and Albrecht, J.~E. 1992.
\newblock Comprehension strategies in the development of a mental model.
\newblock \emph{Journal of experimental psychology: learning, memory, and
  cognition}, 18(4): 777.

\bibitem[{Peng et~al.(2021)Peng, Li, Wiegreffe, and Riedl}]{PengInferring2021}
Peng, X.; Li, S.; Wiegreffe, S.; and Riedl, M.~O. 2021.
\newblock Inferring the Reader: Guiding Automated Story Generation with
  Commonsense Reasoning.
\newblock In \emph{Proceedings of the 3rd Workshop on Narrative Understanding}.

\bibitem[{Purdy et~al.(2018)Purdy, Wang, He, and Riedl}]{purdy2018predicting}
Purdy, C.; Wang, X.; He, L.; and Riedl, M. 2018.
\newblock Predicting generated story quality with quantitative measures.
\newblock In \emph{Proceedings of the AAAI Conference on Artificial
  Intelligence and Interactive Digital Entertainment}, volume~14.

\bibitem[{Radford et~al.(2019)Radford, Wu, Child, Luan, Amodei, and
  Sutskever}]{radford2019language}
Radford, A.; Wu, J.; Child, R.; Luan, D.; Amodei, D.; and Sutskever, I. 2019.
\newblock Language models are unsupervised multitask learners.
\newblock \emph{OpenAI blog}, 1(8): 9.

\bibitem[{Rashkin et~al.(2020)Rashkin, Celikyilmaz, Choi, and
  Gao}]{rashkin2020plotmachines}
Rashkin, H.; Celikyilmaz, A.; Choi, Y.; and Gao, J. 2020.
\newblock PlotMachines: Outline-Conditioned Generation with Dynamic Plot State
  Tracking.
\newblock \emph{arXiv preprint arXiv:2004.14967}.

\bibitem[{Riedl and Young(2010)}]{riedl2010narrative}
Riedl, M.~O.; and Young, R.~M. 2010.
\newblock Narrative planning: Balancing plot and character.
\newblock \emph{Journal of Artificial Intelligence Research}, 39: 217--268.

\bibitem[{Roemmele(2016)}]{roemmele2016writing}
Roemmele, M. 2016.
\newblock Writing stories with help from recurrent neural networks.
\newblock In \emph{Proceedings of the AAAI Conference on Artificial
  Intelligence}.

\bibitem[{Tambwekar et~al.(2018)Tambwekar, Dhuliawala, Martin, Mehta, Harrison,
  and Riedl}]{tambwekar2018controllable}
Tambwekar, P.; Dhuliawala, M.; Martin, L.~J.; Mehta, A.; Harrison, B.; and
  Riedl, M.~O. 2018.
\newblock Controllable Neural Story Plot Generation via Reinforcement Learning.
\newblock \emph{arXiv preprint arXiv:1809.10736}.

\bibitem[{Trabasso and Van Den~Broek(1985)}]{trabasso1985causal}
Trabasso, T.; and Van Den~Broek, P. 1985.
\newblock Causal thinking and the representation of narrative events.
\newblock \emph{Journal of memory and language}, 24(5): 612--630.

\bibitem[{Ware and Siler(2021)}]{ware2021}
Ware, S.; and Siler, C. 2021.
\newblock Sabre: A Narrative Planner Supporting Intention and Deep Theory of
  Mind.
\newblock In \emph{Proceedings of the 17th AAAI International Conference on
  Artificial Intelligence and Interactive Digital Entertainment}.

\bibitem[{Ware and Young(2010)}]{ware2010modeling}
Ware, S.~G.; and Young, R.~M. 2010.
\newblock Modeling Narrative Conflict to Generate Interesting Stories.
\newblock In \emph{AIIDE}.

\bibitem[{Yao et~al.(2019)Yao, Peng, Weischedel, Knight, Zhao, and
  Yan}]{yao2019plan}
Yao, L.; Peng, N.; Weischedel, R.; Knight, K.; Zhao, D.; and Yan, R. 2019.
\newblock Plan-and-write: Towards better automatic storytelling.
\newblock In \emph{Proceedings of the AAAI Conference on AI}.

\end{thebibliography}

\appendix

\section{Appendix}

\subsection{Horizon Regularization}

Repetition penalty accumulates penalty over tokens that the language model has already written. Similarly, n-gram penalties prevent a language model from repeating a string of tokens longer than n. Both of these methods rely on the language model having generated the tokens given the current prompt. Therefore to avoid repetition between prompts, we accumulate error from previous iterations by passing previously generated text as a horizon. That is to say, it can influence the penalties but cannot be used as part of the prompt. Horizon regularization was applied to both backwards BART and EDGAR.

\subsection{Context Documents}

Context was pulled from flash fiction online. \footnote{\url{https://www.flashfictiononline.com}} Scripts for scraping plus the dataset will be released upon publication. Flash fiction online was chosen as the stories are roughly the same length as thought found in the science fiction summary corpus, but require much higher reading comprehension skills. By comparison, the science fiction corpus is only marginally harder to follow than ROC.

\subsection{Experiment details}

The only training that was performed during the completion of this project was that of the ranker and the baseline. No hyper parameter search was performed, all values were the default values provided by Huggingface's Transformer library.

For inference, we chose a high repetition penalty of 10.0 for both bBART and EDGAR. Similarly, a length penalty of 3.0 and max\_length of $150$ was used.

\subsection{Release of code}

We will release our code publicly upon paper decision, but currently it uses internal code and models that are not ready for release as they relate to separate unreleased projects.

\subsection{T/F templates}

\begin{itemize}
    \item $E_j$ depends on $E_i$.
    \item $E_i$ happening after $E_j$ would prevent $E_k$.
    \item Without $E_i$, $E_k$ would have happened before $E_2$.
    \item Is object $O$ mentioned explicitly or implicitly before $E_2$.
    \item $E_j$ contradicts assertions in $E_i$.
    \item $E_i$ depends on $E_j$.
    \item Character $C$ could be removed and the story would still make sense.
    \item $E_i$ could be removed and the story would still make sense.
    \item $E_i,\hdots, E_k$ can be removed and the story would still make sense.
    \item $E_i$ could be removed and the story would still make sense.
    \item $E_i$ changes some property of object $O$/character $C$.
\end{itemize}

where $i < j < k$, $E$ refers to an event, $O$ refers to an object, and $C$ refers to a character.

\subsection{Story Example: EDGAR}

The first bar is tied to the end of a rope running up the side of the tower while the second is tied to a pulley system connected to the elevator shaft.\\
So when Prince Albert climbs down, he uses one of the rails to get to the platform where the elevator shaft connects to the rest of the building.\\
When he comes back down, he takes the ropes and ties them to the ends of the pulley system instead of using the entire length of the staircase.\\
When he went to the top of the tower, he climbed down from the roof onto the steps at the base of the stairs leading to the main stairway.\\
 He then took off his helmet and set himself on the second floor.\\
 From there, he had to climb all the way back down to the bottom of the stairs before he could take off again.\\
 She doesn’t know how to find it, so she can’t even make a phone call to ask for help.\\
 It’s possible that there are people who have access to the tower but they don’t want to tell anyone because they don’t think it’s worth their time or money.\\
 \textbf{There was a princess with long hair locked in a tower far away.\\
 A prince came to save her by climbing her hair and taking her out of the tower.\\}
 
\centerline{\rule{9cm}{0.4pt}}
\bigskip

\noindent If it does not have a pulse then it is dead.\\
The heart rate of an animal tells them if it is alive or dead.\\
There is no such thing as absolute death.\\
He is able to tell the difference between something being alive and something being dead, so when he looks at the house, he feels\\ like he’s seeing things that aren’t there.\\
This makes him feel uncomfortable because he doesn’t want to be in that situation.\\
It’s similar to how people can see ghosts or monsters from inside their head but they don’t know what those things are.\\
\textbf{Hansel’s hand still trembles as he pushes open the twice-cooked door.\\
The last time he saw the house he was glancing back over his shoulder as he and his sister fled into the trees.\\}

\centerline{\rule{9cm}{0.4pt}}
\bigskip    

\noindent The sparrow is watching him.\\
It’s perched on the barrow fence, a tousle of feathers and watchful birds and watch their prey.\\
He’s worried that the sparrow is listening to him.\\
\textbf{Hansel’s hand still trembles as he pushes open the twice-cooked door.\\
The last time he saw the house he was glancing back over his shoulder as he and his sister fled into the trees.\\}
 
\centerline{\rule{9cm}{0.4pt}}
\bigskip    

\noindent “She’s dead.”\\
He doesn’t know how to get out of the house because he isn’t aware of anything else happening.\\
 However, if someone were to come up to him and say “You should go home,” he would have no choice but to run away from them.\\
 He wants to get out of that creepy house.\\
 The only way for him to do this is to push open the door once it’s been cooked.\\
 This would be a terrible thing to do, so he shudders in anticipation of doing it.\\
\textbf{Hansel’s hand still trembles as he pushes open the twice-cooked door.\\
The last time he saw the house he was glancing back over his shoulder as he and his sister fled into the trees.\\}

\centerline{\rule{9cm}{0.4pt}}
\bigskip    

\noindent He doesn’t seem to be able to control his emotions, but that isn’t always the case.\\
 The knife may have been used by him in a momentary panic or just because he felt like pushing through the frosted gingerbread girders on the other side of the door.\\
 It might have been one of those moments where he didn’t know what else to do with his hands.\\
 But then again, there are plenty of people who think they can handle things without any kind of mental breakdown.\\
\textbf{Hansel’s hand still trembles as he pushes open the twice-cooked door.\\
The last time he saw the house he was glancing back over his shoulder as he and his sister fled into the trees.\\}

\centerline{\rule{9cm}{0.4pt}}
\bigskip    

\noindent She would say things like “I’m tired!” or “It’s been a long day!” when she wasn’t really tired.\\
 It’s kind of hard to imagine what being able to do both at once feels like.\\
 The Princess didn't have time to see the world, she had no time to see it at all.\\
 In order to experience the world there needs to be some sort of barrier between herself and everything else.\\
 There are many ways to do this without going too far into detail, but that's the general gist of it.\\
 \textbf{There was a princess with long hair locked in a tower far away. \\
 A prince came to save her by climbing her hair and taking her out of the tower.\\}
 
\centerline{\rule{9cm}{0.4pt}}
\bigskip    
 
 \noindent It made sense that he’d have a strong sense of unease about his mother’s safety.\\
It was a very strange experience for a young child to have, and it made him think about how he’d felt before, which he was very conscious of.\\
He was raised in a home where his mother was the only person who could give a damn.\\
The only thing Hansel had to do was escape.\\
\textbf{Hansel’s hand still trembles as he pushes open the twice-cooked door.\\
The last time he saw the house he was glancing back over his shoulder as he and his sister fled into the trees.\\}

\centerline{\rule{9cm}{0.4pt}}
\bigskip    

\noindent They want a job to provide for them and their family.\\
He has no other option.\\
\textbf{Hansel’s hand still trembles as he pushes open the twice-cooked door.\\
The last time he saw the house he was glancing back over his shoulder as he and his sister fled into the trees.\\}

\centerline{\rule{9cm}{0.4pt}}
\bigskip    

\noindent They just don’t want to admit it.\\
 Because they feel the urge to.\\
 So they go somewhere else.\\
 He has no other option.\\
\textbf{Hansel’s hand still trembles as he pushes open the twice-cooked door.\\
The last time he saw the house he was glancing back over his shoulder as he and his sister fled into the trees.\\}

\centerline{\rule{9cm}{0.4pt}}
\bigskip

\noindent If someone says they're going to use their teeth to open a door, then they probably aren't going to be able to do it very well.\\
But if they say they'll use their fingers instead, then they will be able to get the door open better.\\
And since people have been doing it for so long, it becomes second nature.\\
In order to enter the room, Hansel has to pass through a gate which is much more difficult to open than the main door.\\
 There are many different ways to open the door, but the most common method is to push against the handle with one hand while using the other hand to open the gap between the handle and the door.\\
 This allows Hansel to walk right through the gap without having to put any effort into opening the door.\\
 There’s no reason for him to see it in person.\\
 He doesn’t need to know anything about it other than that it’s there.\\
 The only thing he needs to know is that he can’t live there anymore.\\
\textbf{Hansel's hand still trembles as he pushes open the twice-cooked door.\\
The last time he saw the house he was glancing back over his shoulder as he and his sister fled into the trees.\\}

\subsection{Story Example: Backwards BART}

In this universe, humans are forced to deal with one another by creating monsters that can destroy everything they touch or even kill themselves. It's all about them being human instead of just living on planet Earth.In this universe, Humans are forced to decide between two different worlds by creating monsters which can destroy everything you touch or even die themselves. \\
It’s all about them be human rather than just living on Planet Earth. A battle has been fought between two monstrous creatures who have come. A battle has broken out between two monstrous beings who have come together to fight each other for their freedom. \\
On Earth, a powerful creature called the Nymph emerges from its hiding place in an open forest. A battle has broke out between two huge beings who have came together to fight against each other for Their freedom. \\
On earth, a powerful monster called the Nymph emerges from his hiding place in An open forest. \\
The hero charged at it, both disappearing Into the void. On Earth \textbf{The hero charged at the dragon, both disappearing into the void. The world is saved and people will not forget the sacrifice of the nameless hero.\\}

\centerline{\rule{9cm}{0.4pt}}
\bigskip    

\noindent A little while ago, I got home from work having just finished reading one of my books. As usual, I didn't sleep well.\\
A little bit ago, I get home from work has just finished reading another book. As always, I don't sleep well either. The night Before, I was sleeping peacefully.\\ 
They were eating breakfast together. This morning, I was sitting quietly. The night before -- I was playing Chess with my friends In their apartment.They were eating breakfast. \\
The night before, I was playing chess with my friends in their apartment. They were watching me play chess on TV.The night before, they watched me play chess with my friend in their apartment building. \\
It was very strange to watch them play chess with each other. I woke up feeling like an idiot. I thought it was because of some unknown reason.I woke up feeling as though an alien had come into my apartment building. \\
They were viewing me playing chess with eachother. \textbf{I woke up realizing that I became a cat. I caught a mouse and at the next moment I realized that I'm a human again!\\}

\centerline{\rule{13cm}{0.4pt}}
\bigskip    

\noindent In 1989, two women who have been married since they were 5 years old wrestled together into one team. One man is now famous for his ability to beat men - even if he can't do it all themselves.\\
In 1989, two female wrestlers whom have been married throughout their lives wrestled together onto one team.One man is now known for his abilities to defeat men - even though he can't win them all himself. \\
A young girl has been married throughout both worlds. A A young girl from China was born and brought up on an orphanage whose parents were forced to let her go by their husbands' demands. \\
She grew up watching sports at home with her mother before becoming a wrestler herself.A young girl from Chinese was born and raised on an orphanages whose parents had turned against her fathers' demands.\\
She grew up watching games at home without her mother's demands. She became a wrestler herself after becoming a wrestler himself. A woman broke through- stereotype \\
\textbf{A woman broke through the stereotype of not being able to wrestle. With her father as a strict coach, she won a gold medal for wrestling in India, fulfilling her father's dream.\\}

\centerline{\rule{9cm}{0.4pt}}
\bigskip    

\noindent The house is derelict and crumbling now, but that doesn't mean it isn't still full of ghosts!\\
That’s just not true at all. The fact that he can imagine himself doing it makes him feel more comfortable with what he’s doing\\
If he didn’t enjoy it, he wouldn’t have done it.\\
The first thing is that he had to do it because he was afraid of being kicked out by his brother. And the second thing is that he wanted to see Jemim's room as much as possible before going to bed. So he took a moment to check for any signs of danger.\\
\textbf{He turned out the light and went into Jem's room. He would be there all night, and he would be there when Jem waked up in the morning. But little did Travis know, Jem knew about this all along.\\}

\centerline{\rule{9cm}{0.4pt}}
\bigskip    

\noindent A woman named Stella Smith comes into their house in Puddle Creek on New Year's Eve 2012. They have just had lunch together.\\
A woman named StellaSmith comes into their home in Puddle Rapids on New Year's Eve 2012. We have just having lunch together. \\
The peanut farmer will be selling peanuts for About three years now which he wants her willing to eat some of his goods as well.We have just had breakfast together. The Peanut farmer has already sold peanuts for. The peanut farmer has been selling peanuts for about three years now. \\
He wants her to buy some of his own food as soon as she can.The peanut farmer has started selling peanuts for around three years now.. he wants her to purchase some of his other food as well. \\
Stella loves peanuts because she knows what kind of animal they are eating.Stella loves peppers for about three weeks now. He want her to buy a few of his own foods as soon as possible. Stella loves. \\
\textbf{Stella loves peanuts! She goes to the market every day, and is best friends with the owner.\\}

\centerline{\rule{9cm}{0.4pt}}
\bigskip    

\noindent A young boy named Hansel has been walking through town recently looking at some strange things that are going on outside their walls. In fact, they were just playing cards before anyone noticed them.\\
A young boy called Hansel has Been walking through town lately looking at some odd things that are gone missing outside their walls -- maybe even someone noticing them. In reality, they were probably playing cards before everyone noticed them. He wants to learn more about it though, but can always tell for sure. He wants to know more about it, but can never tell for sure.\\
After all, this is a city of humans who have no interest in living here.He wants to know what happened to him, but can only tell for sure if this is a City of Humans who have little interest in living there. Hansel's hand still feels like an old man with no interest whatsoever.  His mind wanders frantically around trying to figure out how to live here. \\
\textbf{Hansel’s hand still trembles as he pushes open the twice-cooked door. The last time he saw the house he was glancing back over his shoulder as he and his sister fled into the trees.\\}

\centerline{\rule{9cm}{0.4pt}}
\bigskip    

In another world, there were many stories about humans who have lost their lives while trying to survive on this planet. \\
One man told him that they all know each other's names.In another universe, there were several stories about humans Who have lost their life while trying to escape on this continent. One man tells him that they both know each other's names. \\
The story end when an old women had died from dehydration, leaving only his dead body which is not reaching without her help. The story ended when an old woman had died from starvation, leaving only his dead body which she could not reach for. She tried to hide it but couldn't find it.\\
The story ends when an old lady had died from hunger, leaving only His dead body which he could not reach without her help. She tried hiding it herself but couldn't get it anywhere else. \\
There was a Princess with long hair tied in a tower near her. A prince brought to save her through climbing her. \textbf{ There was a princess with long hair locked in a tower far away. A prince came to save her by climbing her hair and taking her out of the tower.\\}

\centerline{\rule{9cm}{0.4pt}}
\bigskip    

\noindent .It was late December 22nd -- another day of quiet for them to enjoy each other's company. The two men0 sat down by the fireplace drinking hot cocoa after dinner.It was late December 23rd -- another day during quiet for us to enjoy each others company. The pair sat down by one firelight after dinner. \\
A man named Randy had just finished an conversation with his neighbor Janice on New Year's Eve. They'd been having breakfast together since midnight.They sat down by A man named Travis had just finished a conversation with his wife Janice on Christmas Eve.\\
 They were having breakfast together at her house.A man named Travis has just finished a discussion with his wifeJanice had just finished talking with his wife Jenice on Christmas eve. \\
They were eating breakfast together at their house. He turned out the lights and went into Jake's room. There would be no night sleep, and he wouldn't be there when they waked up next morning. \\
\textbf{He turned out the light and went into Jem's room. He would be there all night, and he would be there when Jem waked up in the morning. But little did Travis know, Jem knew about this all along.}\\

\centerline{\rule{9cm}{0.4pt}}
\bigskip    

\noindent  In this universe, humans are forced to deal with one another by creating monsters that can destroy everything they touch or even kill themselves.\\
It's all about them being human instead of just living on planet Earth.In this universe, Humans are forced to decide between two different worlds by creating monsters which can destroy everything you touch or even die themselves.\\
It’s all about them be human rather than just living on Planet Earth. A battle has been fought between two monstrous creatures who have come.\\ A battle has broken out between two monstrous beings who have come together to fight each other for their freedom. \\
On Earth, a powerful creature called the Nymph emerges from its hiding place in an open forest. \\
A battle has broke out between two huge beings who have came together to fight against each other for Their freedom. \\
On earth, a powerful monster called the Nymph emerges from his hiding place in An open forest. \\
The hero charged at it, both disappearing Into the void.On Earth \textbf{The hero charged at the dragon, both disappearing into the void. The world is saved and people will not forget the sacrifice of the nameless hero.}\\

\centerline{\rule{9cm}{0.4pt}}
\bigskip    

\noindent The last time I saw my grandfather before leaving home had been during an argument between two people over money. His parents were very angry with him because they didn't want him to leave them alone.\\
The last time I see my grandfather before going away had been during another argument between two men over money. My parents were very upset with him because we didn't want anyone else to leave them together. \\
A young man called after him is telling me how much better their relationship was than theirs. A young man named after him is trying to kill his father for being too strong of a woman. \\
He has no idea who he is or what he really does and that he doesn't like about it. A young man named After him is attempting to kill his dad for being too weak of a woman's strength and that he shouldn't like about this.\\
He has nothing knows who he really is and that he don't like about anything. An evil queen wanted at least one thing from 
\textbf{An evil queen wanted to take revenge on the prettiest girl in the land. She gave her a poison apple, but the girl was saved by a prince.\\}

\subsection{Story Example: ROC}

I got Charlie Horse when I was four years old.\\
He's a brown stuffed horse, and at 35 I still sleep with him at night.\\
He was my best friend, and always laid at the head of my bed.\\
I laid him next to me, smelling his soft fur every night.\\
I liked to listen to my radio as I fell asleep cuddling him.

\end{document}